%% file: main.tex
\documentclass{article}
\usepackage{iclr2027_conference,times}
\input{math_commands.tex}

\usepackage{graphicx}
\usepackage{booktabs}
\usepackage{multirow}
\usepackage{float}
\usepackage[font=small,skip=2pt]{caption}
\usepackage{hyperref}
\usepackage{url}

\newif\ifeasypublic
\easypublictrue
\ifeasypublic
  \iclrfinalcopy % Disable the anonymous author block and review line numbers.
\fi

\title{Easy-3Dfy: Object-Level Geometric Composition for Single-Image 3D Scene Construction}

\author{%
\begin{minipage}{\dimexpr\textwidth-2\tabcolsep\relax}
\centering
\normalsize
Ziang Zhang\textsuperscript{1,2,*}\quad
Wenyou Yi\textsuperscript{1,*}\quad
Zehan Wang\textsuperscript{1,2,**}\\[3pt]
Jiachen Xu\textsuperscript{2}\quad
Xintong Han\textsuperscript{2}\quad
Chunchao Guo\textsuperscript{2}\quad
Zhou Zhao\textsuperscript{1,\textdagger}\\[5pt]
\normalfont\small
\textsuperscript{1}Zhejiang University\qquad
\textsuperscript{2}Tencent
\end{minipage}%
}

\begin{document}
\maketitle
\ifeasypublic
  % Override the conference header set by \maketitle; remove footer and rules.
  \fancyhf{}
  \renewcommand{\headrulewidth}{0pt}
  \renewcommand{\footrulewidth}{0pt}
  \pagestyle{empty}
  \thispagestyle{empty}
  % Unnumbered first-page footnote; author symbols are specified above.
  \begingroup
    \renewcommand{\thefootnote}{}
    \footnotetext{\raggedright
      \textsuperscript{*}Equal Contribution.\quad
      \textsuperscript{**}Project Leader.\quad
      \textsuperscript{\textdagger}Corresponding author.}
  \endgroup
\fi
% \vspace{-10pt}
\begingroup
    \centering
    \includegraphics[width=\textwidth]{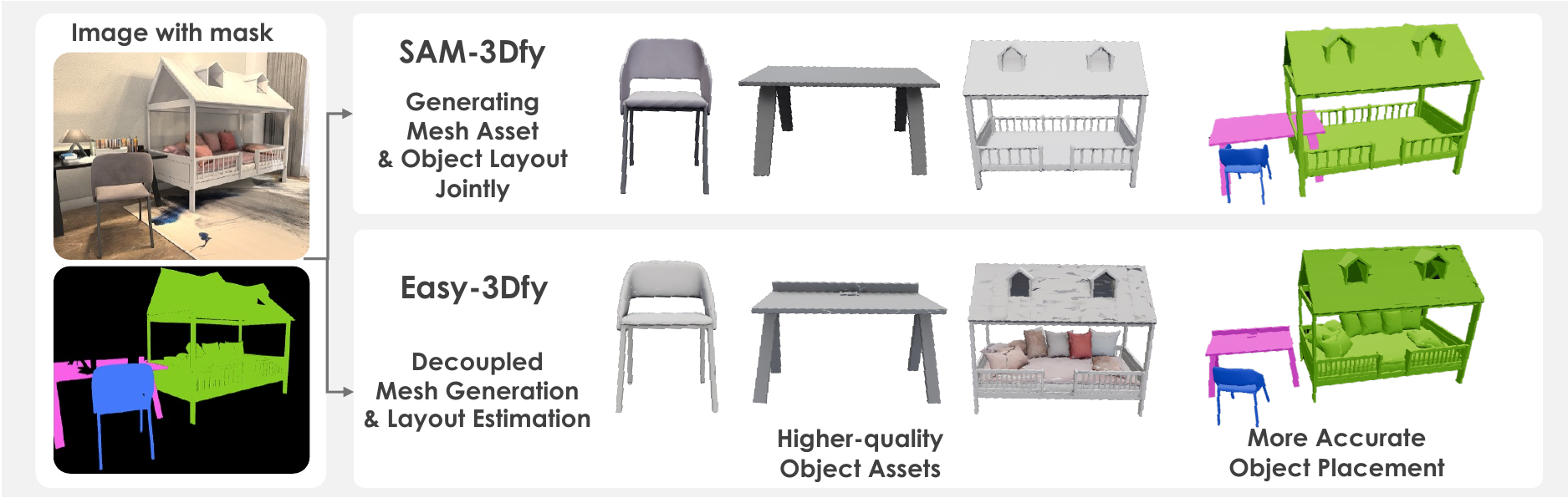}
    \captionof{figure}{\textbf{Qualitative comparison on multi-object scenes.} Given a single image and its object masks, Easy-3Dfy improves object layout and produces more coherent scene assemblies across general examples.}
    \label{fig:teaser}
    \par
\endgroup

\input{sections/abstract}
\input{sections/introduction}
\clearpage
\input{sections/related_work}

\input{sections/method}
\input{sections/experiments}

\input{sections/analysis}
\input{sections/conclusion}
\input{sections/statements}

\bibliography{references}
\bibliographystyle{iclr2027_conference}

\appendix
\input{sections/appendix}

\end{document}

%% file: math_commands.tex
\usepackage{amsmath,amsfonts,bm}

\def\eqref#1{equation~\ref{#1}}
\def\1{\bm{1}}

\DeclareMathAlphabet{\mathsfit}{\encodingdefault}{\sfdefault}{m}{sl}
\SetMathAlphabet{\mathsfit}{bold}{\encodingdefault}{\sfdefault}{bx}{n}

%% file: sections/abstract.tex
% \begin{abstract}
% Single-image 3D scene construction often struggles with generalization due to limited scene-level data, while existing pipelines like SAM-3Dfy lack explicit geometric composition and component-level flexibility. Addressing these limitations, we propose \textbf{Easy-3Dfy}, a compositional workflow that decouples object assets, scene geometry, and layout alignment. Easy-3Dfy reformulates scene construction as an object-level metric layout prediction problem. Specifically, an RPF-based direct layout aligner predicts query-to-anchor correspondence positions from each object-local mesh and its partial pointmap-derived query point cloud, then recovers the local-to-pointmap-scene similarity transform. Trained on object-level alignment pairs constructed through rendering and layout augmentation, this aligner avoids complete-scene supervision. Using identical meshes, Easy-3Dfy outperforms SAM-3Dfy in Scene F-Score and 3D BBox IoU on Scene-Maker and 3D-FUTURE200. Further analysis reveals how assembled scene quality depends on upstream 3D geometry, outlining future directions for single-image 3D scene construction.
% \end{abstract}

\begin{abstract}

Single-image 3D scene construction is a fundamental problem in 3D vision. Recent methods such as SAM-3Dfy enable end-to-end reconstruction of 3D assets and their scene layouts, but still struggle with fine-grained scene geometry and asset quality. Meanwhile, specialized 3D object generators provide detailed object geometry, while 3D foundation models offer strong and generalizable spatial understanding. We propose \textbf{Easy-3Dfy}, a compositional workflow that decouples object generation from scene layout estimation, allowing specialized models to fully exploit their respective strengths. An RPF-based aligner then bridges these capabilities by establishing geometric correspondence between object-local meshes and partial scene pointmaps, recovering object layout for scene composition. Without requiring complete scene-level supervision, the proposed workflow consistently outperforms SAM-3Dfy across multiple geometric metrics on Scene-Maker and 3D-FUTURE200. Our results demonstrate that explicit geometric composition can effectively combine high-quality object generation with strong spatial understanding, while enabling the workflow to benefit independently from advances in both 3D object generation and foundation models.
\ifeasypublic
Project page and code release: \url{https://easy3dfy.github.io}.
\fi

\end{abstract}

%% file: sections/introduction.tex
\section{Introduction}
\label{sec:introduction}

Single-image 3D scene construction is a fundamental problem in 3D vision, aiming to recover a set of complete 3D object assets and arrange them in a common metric coordinate system from a single image. Recent advances have made substantial progress toward this goal. In particular, SAM-3Dfy~\citep{sam3d} provides an end-to-end framework that jointly addresses openset object asset generation and scene layout estimation, demonstrating the feasibility of reconstructing complete 3D scenes from a single image. However, jointly addressing these capabilities within a scene-level pipeline can limit the fine-grained quality of both generated assets and spatial layouts. The object assets produced by such scene-oriented systems can remain less detailed than those from specialized 3D object generation models, while the lack of explicit geometric alignment can lead to interpenetrating and floating object layouts.

\begin{figure*}[t]
    \centering
    \includegraphics[width=0.84\textwidth]{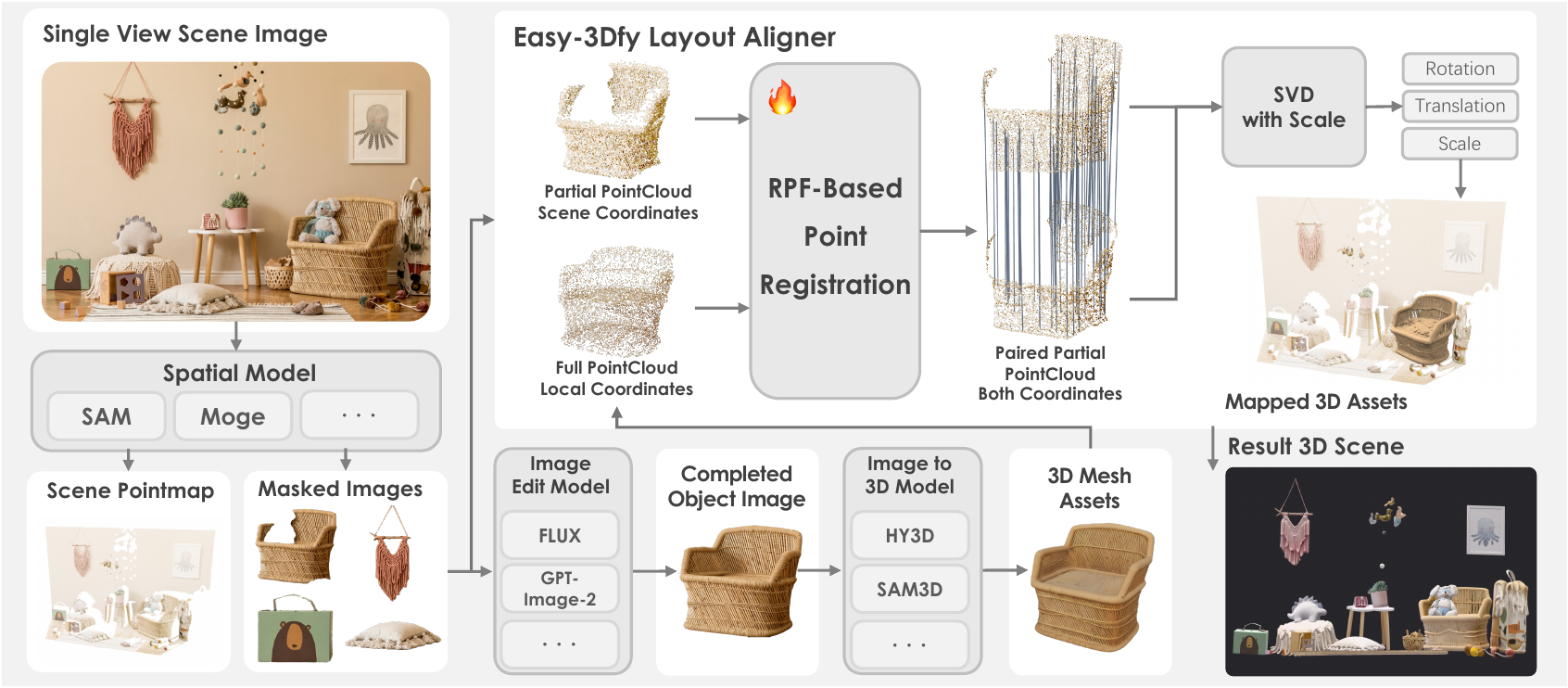}
    \caption{\textbf{Overview of Easy-3Dfy.} Given object masks, a spatial model predicts a scene pointmap. For each object, an image editor produces a completed object image and a replaceable image-to-3D provider generates a local mesh. The RPF-based aligner matches its anchor point cloud to the object-specific partial query extracted from the pointmap, predicts correspondence positions, and recovers a uniform-scale similarity transform via Umeyama fitting. Transforming all complete meshes assembles the final scene.}
    \label{fig:framework}
\end{figure*}

Meanwhile, recent progress in 3D foundation models has provided strong alternatives for the two complementary capabilities involved in scene construction. In particular, 3D object generation models can produce increasingly complete and detailed object assets, while monocular pointmap reconstruction models provide strong scene-level spatial cues. Such decoupling allows each type of model to fully exploit its specialized capability, while making the overall scene construction workflow naturally extensible to future advances in either direction.

Building on these complementary capabilities, we introduce \textbf{Easy-3Dfy}, a compositional workflow that reformulates single-image 3D scene construction as an object-level point cloud registration problem. Specifically, given a complete object asset and the corresponding partial scene point cloud, the goal is to estimate the transformation from the asset-local coordinate system to the scene coordinate system. To address this, Easy-3Dfy incorporates a Rectified Point Flow Model (RPF) aligner to resolve geometric correspondence and recover the relative transform parameters between the two point sets. Specifically, RPF operates as a generative registration model that directly synthesizes surface-aligned point clouds conditioned on the object asset and the scene point cloud, establishing robust point-to-point relationships. This aligner acts as a critical geometric bridge, allowing the independent object generation and scene reconstruction modules to function collaboratively within a unified framework.

To compare and validate the scene geometric construction capability of Easy-3Dfy, we first evaluate the quality of asset composition and scene layout produced by Easy-3Dfy and SAM-3Dfy using the same generated object assets on Scene-Maker~\citep{shi2025scenemaker} and 3D-FUTURE200~\citep{fu2021future}. Easy-3Dfy consistently improves multiple geometric metrics at the scene-level, demonstrating that explicit geometric alignment is effective for high-quality scene construction. Furthermore, by replacing the underlying 3D object generation models, we observe that Easy-3Dfy can further benefit from improvements in upstream object generation, indicating the scalability of the proposed decoupled workflow.

Our contributions are summarized as follows:
\begin{itemize}
\setlength{\itemsep}{0pt}
\setlength{\parsep}{0pt}
\setlength{\parskip}{0pt}
\setlength{\topsep}{2pt}
\item We propose \textbf{Easy-3Dfy}, a compositional workflow for single-image 3D scene construction that decouples object asset generation from scene-level spatial reconstruction, allowing specialized models to fully exploit their respective strengths.
\item We introduce an \textbf{RPF-based geometric aligner} that formulates scene construction as an object-level point cloud registration problem and establishes geometric correspondence between object assets and partial scene point clouds, enabling explicit transformation estimation from the object-local coordinate system to the scene coordinate system.
\item We systematically validate the effectiveness and scalability of compositional scene construction. Controlled comparisons show that explicit geometric composition improves scene geometric quality even with identical object assets, while experiments with different object generation models demonstrate that the proposed workflow can independently benefit from advances in upstream 3D object generation.
\end{itemize}

%% file: sections/related_work.tex
\section{Related Work}
\label{sec:related_work}

\subsection{Single-object image-to-3D generation}
Image-to-3D models provide object-local geometry, texture, and topology that can serve as reusable asset priors. Representative systems improve object-level shape quality, material synthesis, native mesh generation, or generation efficiency~\citep{hunyuan3d2,trellis2,nexus,meshy_t2}. Image-generation models such as FLUX-2, GPT-Image-2, and Gemini Image can further provide completed object images before asset generation~\citep{bfl2025flux2,openai2026gptimage2,google2025nanobanana}. These advances change the role of object generation in scene construction: rather than requiring a scene model to synthesize every asset internally, a compositional system can obtain an object-local mesh from a specialized provider and focus separately on scene placement. In our formulation, these models serve as upstream providers. Easy-3Dfy does not claim to improve their outputs; it discards every provider pose or layout output and estimates a local-to-pointmap-scene transformation from scene geometry.

\subsection{Monocular geometry estimation and point-cloud alignment}
Geometry reconstruction models such as DUSt3R, VGGT, $\pi^3$, Lotus-2, MogeV2 and Depth Anything V3 estimate depth, pointmaps, normals, camera parameters, or ray representations from input images~\citep{wang2024dust3rgeometric3dvision,wang2025vggtvisualgeometrygrounded,ICLR2026_11a09e0a,he2026lotus2advancinggeometricdense,moge2,depthanything3}. These outputs provide scene-level visible geometry, but do not directly define an object's complete layout because of partial visibility, depth uncertainty, and object boundaries. Classical and learned registration methods recover rigid or similarity transforms from correspondences, local features, or iterative refinement~\citep{fischler1981ransac,besl1992icp,qin2022geometric}. However, when registering partial point clouds derived from monocular depth estimates to full point clouds produced by 3D generative models, their effectiveness can be limited by the domain gap between the two geometry sources. Generative registration methods such as RPF and RAP offer a complementary formulation for ambiguous and partial point-cloud alignment~\citep{rpf,rap}. Easy-3Dfy uses an RPF-based direct layout aligner, but shifts its role from generic registration to object-level geometric composition: rendered partial observations and layout augmentation provide alignment supervision without requiring complete calibrated scenes for the alignment stage. Rather than seeking a new general-purpose registration backbone, we train a registration module specialized for single-image 3D scene construction, matching partial query point clouds extracted from monocular pointmaps to object-local anchor geometry.

\subsection{Single-image 3D scene generation and layout prediction paradigms}
Existing single-view scene-generation systems combine object assets and layouts in different ways. Joint methods generate multiple assets and relative poses in one end-to-end process, thereby exploiting global image context while coupling asset generation, layout prediction, and scene-level supervision. Two-stage or post-hoc systems instead assemble objects after asset generation using geometric, physical, differentiable render-and-compare, or relation-aware constraints. CAST, 3D-RE-GEN, MIDI, SceneGen, and SAM-3Dfy exemplify distinct points in this design space~\citep{cast,sautter2025regen,midi,scenegen,sam3d}. These systems commonly operate under an object-conditioned protocol with automatic, user-provided, or precomputed masks. Scene-Maker first decouples the de-occlusion model from 3D object generation; however, it still relies on scene-level supervision to enhance layout prediction performance~\citep{shi2025scenemaker}. Easy-3Dfy more strictly decouples object-asset generation from geometric alignment: it retains reusable asset and geometry priors, discards provider layout outputs, and learns the remaining local-to-pointmap-scene composition directly in the point-cloud domain.

%% file: sections/method.tex
\section{Method}
\label{sec:method}

\subsection{Problem formulation}
\label{sec:formulation}
Given an input image $I$ and $N$ object masks $\{M_i\}_{i=1}^{N}$, Easy-3Dfy predicts a pixel-aligned scene pointmap $P\in\mathbb{R}^{H\times W\times 3}$ and surface-normal map $\mathbf{N}\in\mathbb{R}^{H\times W\times 3}$. Here, $M_i$ denotes the visible region of object $i$; mask estimation is an upstream component and is not studied in this work. Applying $M_i$ to $I$, $P$, and $\mathbf{N}$ yields the object image $I_i$, object-specific partial query point cloud $\mathcal{P}_i$, and its surface normals $\mathcal{N}_i$. Easy-3Dfy converts $I_i$ into an object-local mesh asset $\mathcal{X}_i$ and uses the pointmap scene coordinate system as the common reference frame.

For each pair $(\mathcal{X}_i,\mathcal{P}_i)$, Easy-3Dfy predicts a transform $T_i$ that maps the complete object-local mesh to a position consistent with the geometric surface represented by $\mathcal{P}_i$. We parameterize $T_i$ as a uniform-scale similarity transform:
\begin{equation}
T_i(\mathbf{x}) = s_i \mathbf{R}_i\mathbf{x} + \mathbf{t}_i,
\label{eq:local_to_scene}
\end{equation}
where $s_i>0$ is a uniform scale, $\mathbf{R}_i\in\mathrm{SO}(3)$ is a rotation matrix, and $\mathbf{t}_i\in\mathbb{R}^{3}$ is a translation in the pointmap scene coordinate system. Applying $T_i$ to each complete object-local mesh yields the assembled scene.

Easy-3Dfy is connected by three explicit interfaces: object-local meshes, a scene pointmap, and local-to-pointmap-scene transforms. These interfaces make the geometric composition stage explicit, separating reusable asset and scene-geometry modules from placement estimation while allowing each module to be replaced without redefining the composition problem.

\subsection{Constructing object assets from a single image}
\label{sec:asset_construction}
As illustrated in Fig.~\ref{fig:framework}, Easy-3Dfy first converts the input image into object-local mesh assets. For a specific object, the following three-stage workflow constructs its asset.

\paragraph{Object extraction.}
Object-mask estimation is outside the scope of this work. Easy-3Dfy accepts masks from an upstream automatic or user-prompted segmentation model, such as SAM~\citep{kirillov2023segment}, and uses each $M_i$ to define the visible object region and crop box. In our benchmark experiments, the masks are provided by the datasets as shared inputs for consistent comparison across scene-construction methods.

\paragraph{Occlusion-aware completion.}
For each object, a prompted image-editing model completes occluded regions. Its reference inputs include the original image, the object crop, and the crop with the object masked. These inputs provide global image context, information about neighboring occluders, and the object's outline, facilitating recovery of occluded appearance and shape cues.

\paragraph{Image-to-3D asset generation.}
Finally, an image-to-3D generative model generates the complete object-local mesh asset $\mathcal{X}_i$ conditioned on the completed object image. 

By applying this workflow to each object to be constructed, Easy-3Dfy produces the mesh-asset set $\{\mathcal{X}_i\}_{i=1}^{N}$. We use each module according to its original inference protocol; the concrete providers and implementation settings are specified in the appendix.

\subsection{Layout estimation and scene assembly}
\label{sec:rpf_alignment}
\paragraph{Object-specific partial query construction.}
Following Sec.~\ref{sec:formulation}, $M_i$ selects object $i$ from the scene pointmap and normal map. Rule-based filtering removes boundary artifacts, outliers, and interpolation contamination, yielding the visible object-specific partial query $(\mathcal{P}_i,\mathcal{N}_i)$ in the scene coordinate system. From the complete object-local mesh $\mathcal{X}_i$, we sample the anchor point cloud and normals $(\mathcal{A}_i,\mathcal{N}^{\mathcal{A}}_i)$. These anchor and query pairs form the geometric input to RPF; filtering, normal construction, and sampling details appear in the appendix.

\paragraph{RPF-based aligner and direct registration.}
RPF~\citep{rpf} is a DiT-based conditional generative model whose condition is divided into anchor and query groups. Each point is represented by its XYZ coordinates and surface normal. Given the anchor geometry $(\mathcal{A}_i,\mathcal{N}^{\mathcal{A}}_i)$ and partial query $(\mathcal{P}_i,\mathcal{N}_i)$, RPF predicts, for every query point $\mathbf{p}_{ij}\in\mathcal{P}_i$, a correspondence position $\widehat{\mathbf{a}}_{ij}$ in the anchor coordinate system. Intuitively, if $\mathbf{p}_{ij}$ represents a visible headlamp in the input image, $\widehat{\mathbf{a}}_{ij}$ localizes the matching headlamp region on the complete car asset. The predicted pairs $\{(\mathbf{p}_{ij},\widehat{\mathbf{a}}_{ij})\}_j$ therefore establish pointwise query-to-anchor correspondences. Figure~\ref{fig:rpf_training_pipeline} illustrates how this correspondence predictor is trained from object-level data.

The RPF condition deliberately excludes RGB and texture. Appearance can vary substantially across illumination, rendering, and generated assets, whereas XYZ coordinates and surface normals provide a more direct geometric signal for layout alignment. Consequently, the aligner operates only on six-dimensional XYZ--normal features and performs geometry-only registration. We further examine the effect of RGB conditioning in Appendix~\ref{sec:appendix_additional_experiments}.

\paragraph{Pose recovery and scene assembly.}
RPF pairs each query point $\mathbf{p}_{ij}$ with a predicted correspondence position $\widehat{\mathbf{a}}_{ij}$ on the anchor geometry. We apply Umeyama fitting to these correspondences and compose the fitted transform with the preprocessing transforms to recover the local-to-pointmap-scene transform $T_i$ in Eq.~\ref{eq:local_to_scene}. The point-set normalization, fitting procedure, and numerical safeguards are detailed in the appendix. Finally, we apply $T_i$ to the complete object-local mesh, rather than only to its sampled anchors, and assemble all transformed meshes into the output scene.

\subsection{Training the RPF-based aligner}
\label{sec:rpf_training}
\paragraph{RPF formulation.}
We follow the original RPF architecture and training formulation. Conditioned on the anchor and partial query geometry $(\mathcal{A}_i,\mathcal{N}^{\mathcal{A}}_i,\mathcal{P}_i,\mathcal{N}_i)$, RPF predicts a velocity field that transports a noisy point configuration to the query points' correspondence positions on the complete object. Let $\mathbf{x}_0$ denote these ground-truth correspondence positions, $\mathbf{x}_1\sim\mathcal{N}(\mathbf{0},\mathbf{I})$ denote Gaussian noise, and $t$ denote a sampled flow time. The rectified-flow path and its target velocity are
\begin{equation}
\mathbf{x}_t=(1-t)\mathbf{x}_0+t\mathbf{x}_1,\qquad
\mathbf{v}_t=\mathbf{x}_1-\mathbf{x}_0,
\label{eq:rpf_interpolation}
\end{equation}
and the model minimizes
\begin{equation}
\mathcal{L}_{\mathrm{RPF}}
=\operatorname{MSE}\!\left(
\mathbf{v}_\theta(\mathbf{x}_t,t\mid
\mathcal{A}_i,\mathcal{N}^{\mathcal{A}}_i,\mathcal{P}_i,\mathcal{N}_i),
\mathbf{x}_1-\mathbf{x}_0
\right),
\label{eq:rpf_loss}
\end{equation}
where $\mathbf{v}_\theta$ is the time-dependent velocity field predicted by the DiT parameterized by $\theta$.

\begin{figure*}[t]
\centering
\includegraphics[width=\textwidth]{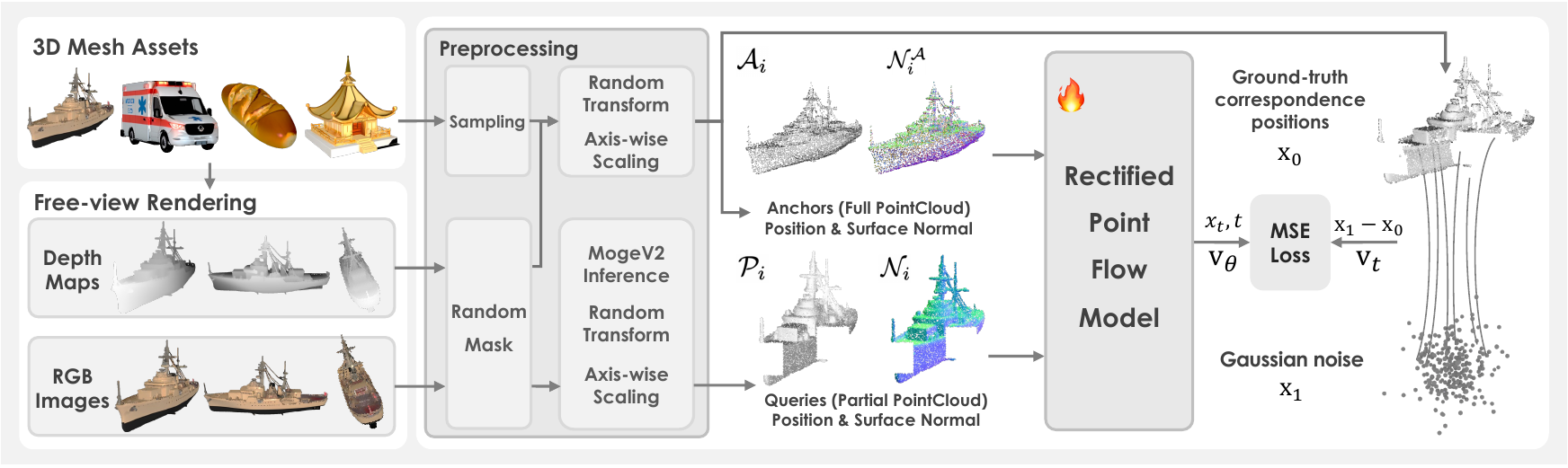}
\caption{\textbf{Object-level data construction and RPF training.} Aligned RGB--depth renders yield a monocular partial query and pixel-matched ground-truth correspondence positions. Independent anchor- and query-side geometric augmentations define the registration task, on which RPF learns the conditional velocity field.}
\label{fig:rpf_training_pipeline}
\end{figure*}

\paragraph{Object-level alignment-pair construction.}
As illustrated in Fig.~\ref{fig:rpf_training_pipeline}, each training sample starts from a single object mesh. We sample an anchor point cloud $\mathcal{A}_i$ and its surface normals $\mathcal{N}^{\mathcal{A}}_i$ from the mesh surface, and render the object from randomly sampled viewpoints to obtain pixel-aligned RGB and depth images. We apply the same monocular geometry model used by Easy-3Dfy to each rendered RGB image, obtaining the partial query point cloud $\mathcal{P}_i$ and surface normals $\mathcal{N}_i$. Random occlusions are then applied to the partial query to simulate the incomplete geometry encountered at inference time.

The ground-truth correspondence positions $\mathbf{x}_0$ are recovered from the rendered depth image at the retained pixels. Because the RGB and depth renders are pixel-aligned, each query point in $(\mathcal{P}_i,\mathcal{N}_i)$ is paired with one position in $\mathbf{x}_0$ on the complete object surface. To expose RPF to changes of coordinate frame and object pose, we divide the training variables into a query-side group $(\mathcal{P}_i,\mathcal{N}_i)$ and an anchor-side group $(\mathcal{A}_i,\mathcal{N}^{\mathcal{A}}_i,\mathbf{x}_0)$, and apply independent random rotations, scales, and translations to the two groups. We additionally apply random axis-wise scaling to perturb object shape, improving robustness to geometric variation across mesh providers. The appendix provides the detailed rendering, occlusion, sampling, and augmentation settings.

Unlike scene-level 3D generation pipelines, Easy-3Dfy learns point-cloud registration at the object level. Its training data therefore require neither complete scene-level data nor explicit modeling of inter-object relationships during data construction. This object-level formulation substantially simplifies training-pair synthesis and allows the dataset to be scaled efficiently.

\subsection{Implementation summary}
\label{sec:implementation}
Our implementation instantiates each interface with a set of interchangeable foundation models. For monocular scene geometry, we use MogeV2~\citep{moge2} and Depth Anything V3~\citep{depthanything3}. MogeV2 directly predicts the surface-normal map together with scene geometry, whereas for Depth Anything V3 we estimate surface normals from its predicted pointmap. The partial geometric observations used to train RPF are constructed with MogeV2. The training meshes are drawn from the publicly released synthetic 3D asset collection introduced by Orient Anything V2~\citep{orientanythingv2}, which was created through its generative 3D asset pipeline.

The main comparison uses MogeV2 for scene geometry and SAM-3Dfy~\citep{sam3d} meshes as object assets. In the modular provider studies, occlusion-aware completion uses FLUX-2-pro~\citep{bfl2025flux2} and GPT-Image-2~\citep{openai2026gptimage2}, while object-local meshes are generated with Hunyuan3D-2.0 and Hunyuan3D-3.1~\citep{hunyuan3d2,tencent2026hy3d31}, TRELLIS-2.0~\citep{trellis2}, Meshy-T2~\citep{meshy_t2}, and Tripo-V3~\citep{tripo2026v3}. All mesh providers are used only to supply object-local assets: Easy-3Dfy does not consume provider-specific pose or layout predictions.

The appendix reports the completion prompt, sampling rules, RPF training and inference settings, geometric augmentation, similarity fitting, and evaluation-time alignment.

%% file: sections/experiments.tex
\section{Experiments}
\label{sec:experiments}

\subsection{Experimental setup}
\label{sec:setup}
\paragraph{Datasets.} We evaluate on 1,000 scenes sampled from Scene-Maker~\citep{shi2025scenemaker} and 200 scenes sampled from the 3D-FUTURE~\citep{fu2021future} test set, denoted 3D-FUTURE200. Scene-Maker covers broader, more open-set object categories, whereas 3D-FUTURE200 primarily contains indoor furniture scenes. Because mask estimation is outside our scope, all methods receive the object masks provided by each dataset, isolating scene construction from segmentation quality and ensuring consistent comparison.

\paragraph{Compared methods.} We compare Easy-3Dfy with SAM-3Dfy~\citep{sam3d}, MIDI~\citep{midi}, and SceneGen~\citep{scenegen}. In the main Easy-3Dfy setting, we use the SAM-3Dfy meshes as object assets and estimate their layouts from the mesh and pointmap geometry. We further evaluate the effect of the RPF design, mesh provider, and pointmap and occlusion-aware completion models through the ablations below.

\paragraph{Metrics.} We report Scene-CD, Scene-FScore, Object-CD, Object-FScore, and scene-aligned per-object 3D bounding-box IoU. Chamfer Distance (CD) is the sum of the two non-squared mean nearest-neighbor distances between sampled point sets~\citep{fan2017pointset}. F-score is the harmonic mean of nearest-neighbor precision and recall under a distance threshold $\tau$~\citep{knapitsch2017tanks}; we use $\tau=0.1$, following the scene-level threshold used by MIDI~\citep{midi}. We use the same definitions for scene- and object-level point sets. The evaluation-time scene and object alignment procedures, point-sampling settings, averaging rules, and the interpretation of object-level diagnostics are common evaluation details and are specified in the appendix.

\subsection{Main results}
\label{sec:main_results}

Table~\ref{tab:main_results} reports the results on Scene-Maker and 3D-FUTURE200. On both datasets, Easy-3Dfy consistently outperforms MIDI and SceneGen on all reported metrics. The gains in both scene-level geometric fidelity and BBox IoU demonstrate more accurate scene layouts, while the improved Object-CD and Object-FScore further indicate higher-quality object meshes.

We next compare Easy-3Dfy with SAM-3Dfy. In the main setting, Easy-3Dfy uses the same object meshes produced by SAM-3Dfy; consequently, the two methods have identical Object-CD and Object-FScore, isolating the effect of layout prediction. Easy-3Dfy achieves better Scene-CD, Scene-FScore, and BBox IoU on both datasets, obtaining the best layout metrics among the compared methods under the shared-mask evaluation protocol. This result shows that direct geometric alignment can improve object placement without jointly training asset generation and layout prediction in an end-to-end scene generator. We evaluate additional mesh providers and the GT Mesh ceiling oracle separately in the ablation study.

\begin{table*}[t]
\centering
\small
\setlength{\tabcolsep}{5pt}
\renewcommand{\arraystretch}{0.84}
\caption{Main results on Scene-Maker and 3D-FUTURE200. Easy-3Dfy and SAM-3Dfy share identical meshes, isolating layout prediction.  Bold marks the best result per dataset.}
\label{tab:main_results}
\begin{tabular}{llccccc}
\toprule
\multirow{2}{*}{Dataset} &
\multirow{2}{*}{Method} &
\multicolumn{2}{c}{Scene-level} &
\multicolumn{3}{c}{Object-level} \\
\cmidrule(lr){3-4}\cmidrule(l){5-7}
& &
CD$\downarrow$ & F-Score$\uparrow$ &
CD$\downarrow$ & F-Score$\uparrow$ & BBox IoU$\uparrow$ \\
\midrule
\multirow{4}{*}{Scene-Maker}
& MIDI      & 0.442 & 0.329 & 0.379 & 0.384 & 0.065 \\
& SceneGen  & 0.455 & 0.368 & 0.264 & 0.543 & 0.085 \\
& SAM-3Dfy  & 0.261 & 0.615 & \textbf{0.188} & \textbf{0.695} & 0.366 \\
& \textbf{Easy-3Dfy} & \textbf{0.242} & \textbf{0.649} & \textbf{0.188} & \textbf{0.695} & \textbf{0.399} \\
\midrule
\multirow{4}{*}{3D-FUTURE200}
& MIDI      & 0.275 & 0.627 & 0.171 & 0.736 & 0.319 \\
& SceneGen  & 0.320 & 0.584 & 0.168 & 0.747 & 0.292 \\
& SAM-3Dfy  & 0.146 & 0.820 & \textbf{0.139} & \textbf{0.807} & 0.517 \\
& \textbf{Easy-3Dfy} & \textbf{0.140} & \textbf{0.836} & \textbf{0.139} & \textbf{0.807} & \textbf{0.554} \\
\bottomrule
\end{tabular}
% \vspace{-2mm}
\end{table*}

% TODO: Insert Fig. 4 with matched qualitative comparisons among SAM-3Dfy,
% Easy-3Dfy, MIDI, and SceneGen. Use matched camera, scene, and rendering rules.

\begin{figure*}[t]
\centering
\includegraphics[width=\textwidth]{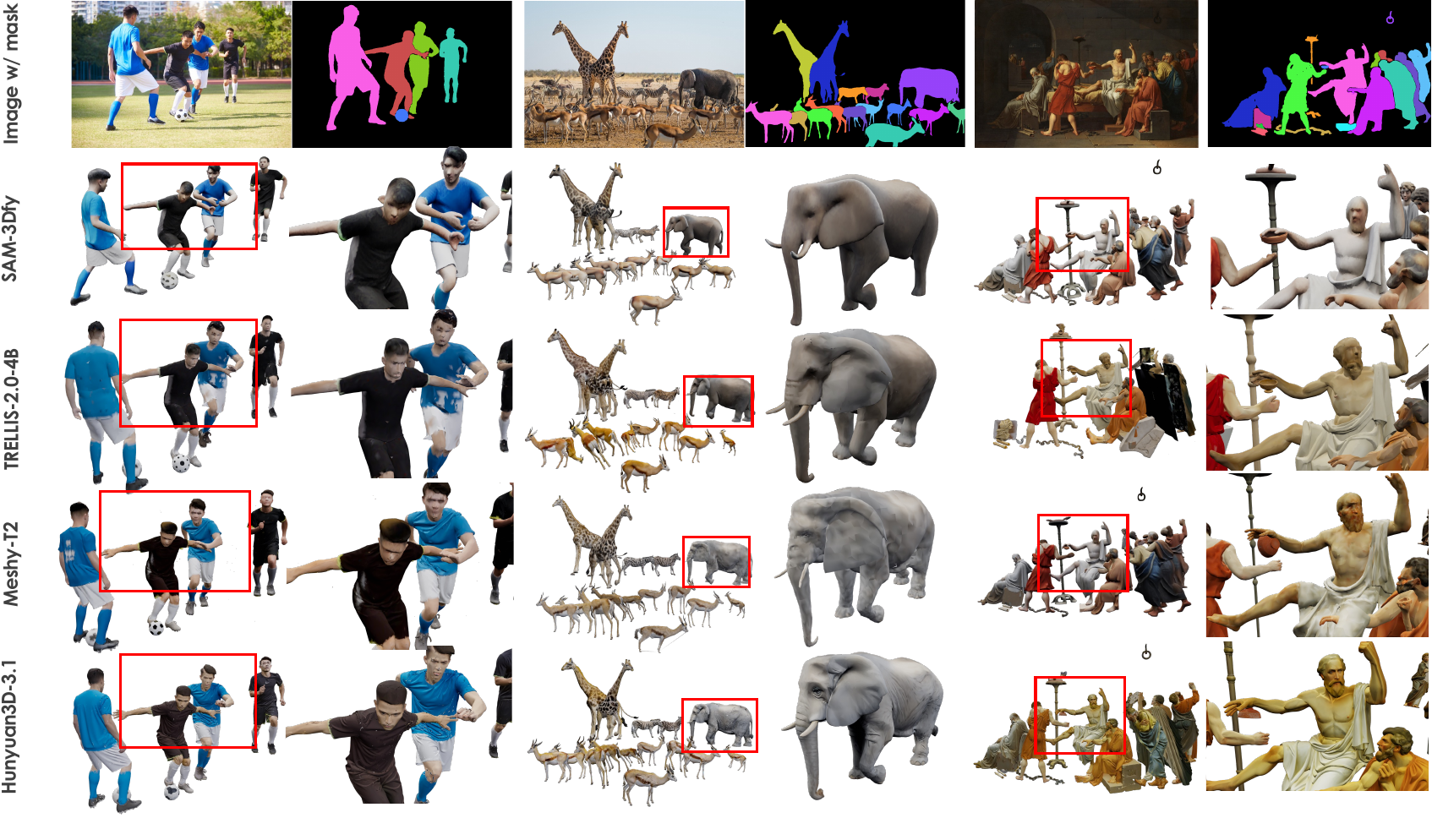}
\caption{\textbf{How mesh quality affects visual and layout quality.} Matched input--panorama--close-up comparisons. Each case consists of two columns, with the first showing the reconstructed full 3D scene and the second showing close-up views of the same asset regions. }
\label{fig:metric_interpretation}
\end{figure*}

\subsection{Ablations}
\label{sec:ablations}
We ablate three components that determine the final constructed scene: the geometric preprocessing and input channels of RPF, the pointmap and occlusion-aware completion models, and the object mesh provider. These studies respectively measure sensitivity to the design of the alignment module , scene-geometry estimation and object completion, and asset quality.

\paragraph{Alignment design.} We further ablate point-cloud preprocessing and RPF input channels. Rule-based cleaning suppresses boundary and interpolation noise but may also remove useful geometry. As RPF is trained on augmented data and tolerates moderate noise, omitting cleaning has only a small effect at the reported precision, primarily a slight decrease in Scene-FScore. Removing surface-normal conditioning degrades all three layout metrics. Full results appear in Appendix~\ref{sec:appendix_additional_experiments}.

\paragraph{Scene geometry and object completion.} Beyond the mesh provider, we compare different pointmap and occlusion-aware completion models in Table~\ref{tab:provider_replacement}, using Meshy-T2 as the common mesh provider. All reported variants use the six-channel XYZ--normal RPF input. The controlled runs cover MogeV2 with FLUX-2 and GPT-Image-2 on both datasets, as well as DAv3 with FLUX-2. GPT-Image-2 slightly improves the scene-level metrics on 3D-FUTURE200 but does not consistently improve the object-level metrics or the results on Scene-Maker, indicating that the effect of the completion model is dataset- and metric-dependent. Refer to Appendix~\ref{sec:appendix_additional_experiments} for full results.

\vspace{-0.5em}
\paragraph{Mesh-provider comparison.} Figure~\ref{fig:metric_interpretation} compare mesh providers by disentangling geometric compatibility from visual detail. Full-scene views highlight improvements in scale, extent, and placement, while close-ups highlight texture and fine-detail quality. More advanced 3D mesh generators not only produce higher-quality textures, but also recover finer-grained surface geometry, such as the orientation of limbs for articulated objects including humans and animals. Moreover, these models often provide more accurate estimates of the object's extent along the depth direction, resulting in object dimensions and aspect ratios that better match the geometric distribution observed in the pointmap. Such meshes provide more faithful surface cues for correspondence and can therefore yield more accurate transformation estimates during subsequent alignment. We analyzes this distinction quantificationally in Section~\ref{sec:asset_geometry_analysis}, and additional visualization results are provided in the Appendix ~\ref{sup:additional_visualization}.

\begin{table*}[t]
\centering
% ==================== 左侧表格 ====================
\begin{minipage}[t]{0.5\textwidth}
\vspace{0pt}
\centering
\caption{Scene-geometry and object-completion ablation on Scene-Maker with Meshy-T2 and six-channel XYZ--normal RPF input.}
\label{tab:provider_replacement}

% --- 控制参数：在此处微调字号和列间距 ---
\scriptsize                         % 可选: \small, \footnotesize, \scriptsize
\renewcommand{\arraystretch}{1.1}  % 控制行高
\setlength{\tabcolsep}{0.3pt}      % 控制列间距 (数字越大列距越宽)

\begin{tabular}{lccccc}
\toprule
Variant & Sce. CD$\downarrow$ & Sce. F$\uparrow$ & Obj. CD$\downarrow$ & Obj. F$\uparrow$ & BBox IoU$\uparrow$ \\
\midrule
MogeV2 + FLUX-2      & 0.253 & 0.632 & 0.201 & 0.667 & 0.381 \\
MogeV2 + GPT-Image-2 & 0.254 & 0.630 & 0.208 & 0.656 & 0.381 \\
DAv3 + FLUX-2        & 0.267 & 0.609 & 0.215 & 0.640 & 0.356 \\
\bottomrule
\end{tabular}
\end{minipage}% <--- 结尾的 % 极其重要，防止换行
\hfill
% ==================== 右侧表格 ====================
\begin{minipage}[t]{0.47\textwidth}
\vspace{0pt}
\centering
\caption{Alignment-design ablation on Scene-Maker. ``Uncleaned Point'' omits point-cloud cleaning, ``XYZ only'' removes surface-normal channel.}
\label{tab:alignment_design}

% --- 控制参数：在此处微调字号和列间距 ---
\scriptsize                         % 可选: \small, \footnotesize, \scriptsize
\renewcommand{\arraystretch}{1.1}  % 控制行高
\setlength{\tabcolsep}{5.0pt}      % 控制列间距

\begin{tabular}{lccc}
\toprule
Variant & Sce. CD$\downarrow$ & Sce. F$\uparrow$ & BBox IoU$\uparrow$ \\
\midrule
Standard        & \textbf{0.242} & \textbf{0.649} & \textbf{0.399} \\
Uncleaned Point & \textbf{0.242} & 0.646          & \textbf{0.399} \\
XYZ only        & 0.245          & 0.641          & 0.394          \\
\bottomrule
\end{tabular}
\end{minipage}
\end{table*}

%% file: sections/analysis.tex
\begin{figure}[t]
\centering
\includegraphics[width=\textwidth]{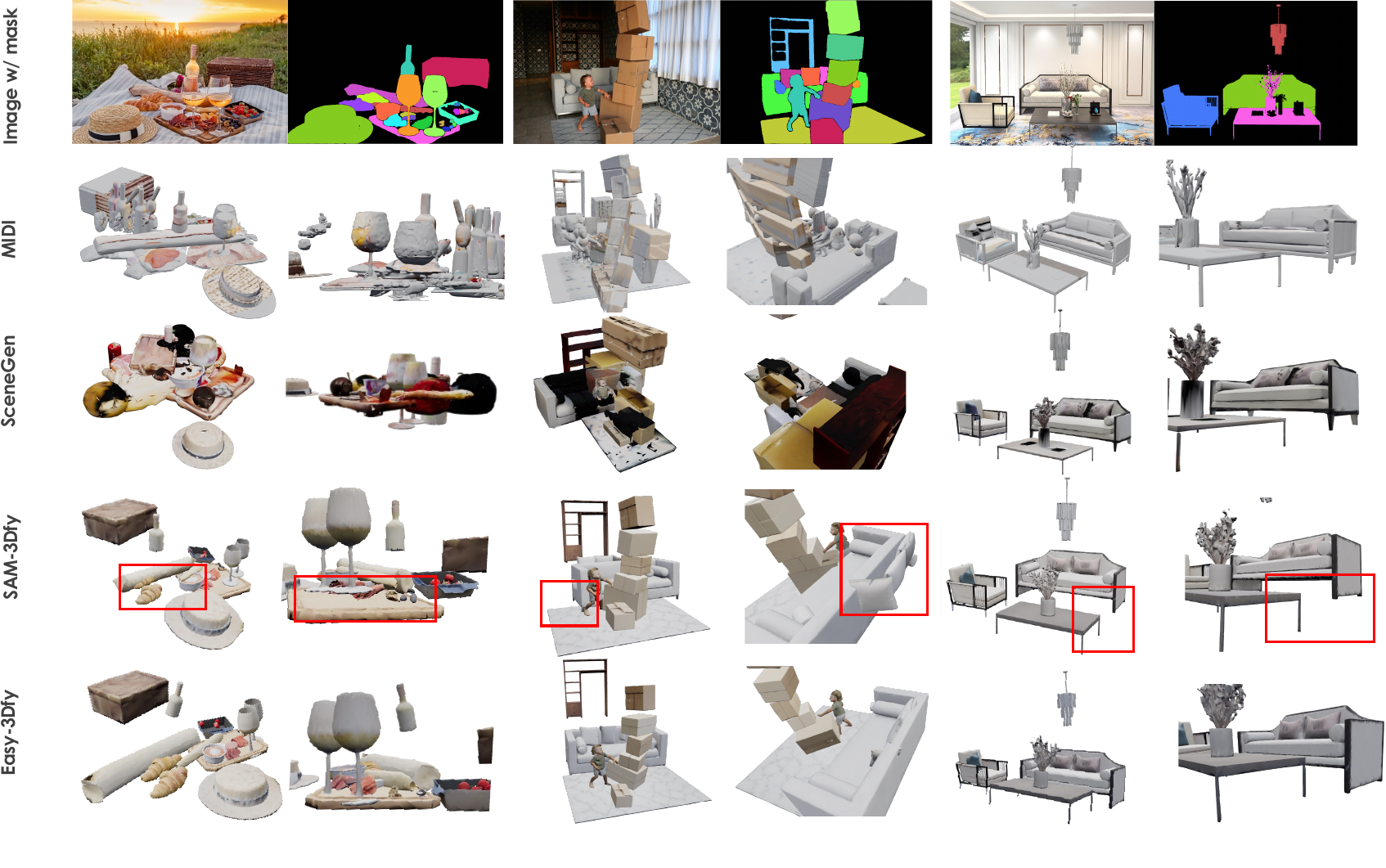}
\caption{\textbf{What direct alignment improves.} Matched comparisons of MIDI, SceneGen, SAM-3Dfy, and Easy-3Dfy under a common rendering protocol. The multi-view results show that Easy-3Dfy reduces floating and interpenetrating objects while improving object ordering and placement.}
\label{fig:direct_alignment_analysis}
\end{figure}

\vspace{-1em}
\section{Analysis and Metric Interpretation}
\label{sec:analysis}

\subsection{What does direct alignment improve?}
\label{sec:direct_alignment_analysis}

Figure~\ref{fig:direct_alignment_analysis} compares these methods with matched inputs and views, exposing differences in the scale, orientation, ordering and relationships of contact of the object. MIDI (trained on 3D-FRONT) and SceneGen (trained on 3D-FUTURE) generally recover the object categories in a multi-object image, but often distort relative orientation, ordering, and spacing in openset. 

Compared with previous methods, SAM-3Dfy is able to predict the approximate positions and ordering of objects in the scene in an open-set setting. However, as shown in Figure~\ref{fig:direct_alignment_analysis}, its predicted layouts still exhibit noticeable geometric inconsistencies. Some objects suffer from interpenetration, such as the pillow floating behind the sofa backrest, while the tabletop and chairs are not properly aligned with the flat ground plane. These artifacts indicate that SAM-3Dfy still struggles to preserve fine-grained spatial relationships and contact relationships between objects. Using the same SAM-3Dfy meshes, Easy-3Dfy instead registers each asset to its pointmap-derived partial query, substantially improving the spatial consistency of object layout and resolving the aforementioned ambiguities of spatial relationships.

Note that small or severely occluded objects, contaminated masks, and unreliable pointmaps for reflective or transparent surfaces can leave the transform underconstrained, producing an inaccurate pose. Appendix~\ref{sec:failure_cases} analysis failure patterns of these cases.

\subsection{Geometric compatibility enhances attainable layout quality}
\label{sec:asset_geometry_analysis}

\begin{table}[H]
\centering
% ==================== 左侧：分析文本 ====================
\begin{minipage}[t]{0.43\textwidth}
\vspace{0pt}
\small
Table~\ref{tab:mesh_source_ablation} compares mesh providers under a fixed pointmap, mask, RPF model, and inference setting. Dataset GT meshes provide an upper-bound reference under the current workflow. The aligner works across all tested providers, demonstrating broad provider generality, while fine-grained layout remains sensitive to asset geometry. Dimensions and proportions that better match the partial observation improve geometric compatibility and facilitate alignment, whereas richer geometric detail or higher-quality texture alone does not guarantee a more accurate layout. This suggests that improvements in 3D asset generation can directly translate into more accurate scene alignment, without requiring any modification or retraining of the Easy-3Dfy aligner. Complete two-dataset results are provided in Appendix~\ref{sec:appendix_additional_experiments}.
\end{minipage}% <--- 极其关键的 %，防止出现额外空格导致超宽换行
\hfill
% ==================== 右侧：表格 ====================
\begin{minipage}[t]{0.55\textwidth}
\vspace{0pt}
\centering
\caption{Mesh-provider comparison on the Scene-Maker dataset with all other Easy-3Dfy components fixed. GT provides an upper-bound reference under the current layout workflow.}
\label{tab:mesh_source_ablation}
\footnotesize
\renewcommand{\arraystretch}{1.3}
\setlength{\tabcolsep}{0pt} % 缩小列间距以适应单侧宽度

\begin{tabular}{lccccc}
\toprule
Mesh & Sce. CD$\downarrow$ & Sce. F$\uparrow$ & Obj. CD$\downarrow$ & Obj. F$\uparrow$ & BBox IoU$\uparrow$ \\
\midrule
GT          & \textbf{0.242} & \textbf{0.660} & \textbf{0.144} & \textbf{0.789} & \textbf{0.415} \\
SAM-3Dfy    & 0.242 & 0.649          & 0.188          & 0.695          & 0.399          \\
TRELLIS-2.0 & 0.284          & 0.557          & 0.273          & 0.548          & 0.317          \\
HY3D-2.0    & 0.295          & 0.523          & 0.278          & 0.514          & 0.291          \\
HY3D-3.1    & 0.256          & 0.629          & 0.196          & 0.678          & 0.385          \\
Meshy-T2    & 0.253          & 0.632          & 0.201          & 0.667          & 0.381          \\
Tripo-V3    & 0.255          & 0.632          & 0.194          & 0.678          & 0.391          \\
\bottomrule
\end{tabular}
\end{minipage}
\end{table}

\subsection{What geometry-centric layout metrics omit}
\label{sec:metric_interpretation}
The preceding ablations and analyses demonstrate both the effectiveness and the scope of the current evaluation. Scene-CD, Scene-FScore, and scene-aligned 3D bounding-box IoU improve with asset quality and registration accuracy, showing that they effectively measure geometric fidelity and spatial arrangement. However, they do not directly reflect appearance fidelity, texture and material quality, semantic identity, scene completeness, or perceptual plausibility. Even within geometry, point-set distances and bounding-box overlap have limited sensitivity to topology and fine surface structure.

The close-ups in Figure~\ref{fig:metric_interpretation} and Figure~\ref{fig:more_vis} make this boundary explicit: a scene can achieve a better layout score while retaining blurred textures or local details, whereas a visually detailed asset can remain geometrically incompatible with the observed scene. Future evaluation protocols should therefore combine geometric, visual, semantic, and completeness criteria to provide a more complete measure of generated-scene quality. 

%% file: sections/conclusion.tex
\section{Conclusion}
\label{sec:conclusion}

We presented Easy-3Dfy, a modular framework that formulates single-image 3D scene construction as object-level geometric composition across replaceable object assets, scene geometry, and metric layout alignment. To address the core layout-alignment problem, we train an RPF aligner on scalable object-level data to recover local-to-scene transforms, improving layout accuracy without requiring complete scene supervision. Experiments demonstrate consistent gains under the shared-mask protocol and applicability across mesh providers. We also identify limitations of geometry-centric metrics and motivate more complete evaluation protocols for generated 3D scenes.

%% file: sections/statements.tex
\section*{AI use statement}
All research ideas, problem formulation, and experimental design were developed by the human researchers. We used AI-assisted programming tools to support code development and large language models to polish the manuscript's language. All code, manuscript content, and cited references were carefully reviewed and verified by the authors, who take full responsibility for the final work.

\section*{Ethics statement}
Based on our review, the datasets and visual materials used in this work do not raise identifiable ethical, security, copyright, or privacy concerns. We use these resources solely for research purposes and follow their applicable licenses and terms of use.

\section*{Reproducibility statement}
The appendix documents the implementation and evaluation details needed to reproduce our method. The supplementary material will include an inference example implemented entirely with open-source models. We commit to publicly releasing our method in the future.

%% file: sections/appendix.tex
\newpage
\section{Appendix}
\label{sec:appendix}
\setcounter{figure}{0}
\renewcommand{\thefigure}{A\arabic{figure}}
\renewcommand{\theHfigure}{A.\arabic{figure}}

\subsection{Additional implementation details}
\label{sec:appendix_implementation}

\paragraph{Occlusion-aware completion prompt.}
The completion model receives three reference images. We use the following prompt:

\begin{quote}
Complete the redrawing and restoration of the target object using the three provided reference images:
\begin{enumerate}
    \item The first image is the original cropped region, containing the target object together with its local foreground and background.
    \item The second image is the crop-aligned target mask. White denotes the target, and black denotes non-target regions.
    \item The third image is the complete scene image, which provides global context and a reference for the target's actual appearance in the scene.
\end{enumerate}
Use these inputs to extract the target object precisely and restore it intelligently:
\begin{enumerate}
    \item \textbf{Structural completion and de-occlusion:} Completely ignore and remove any foreground objects occluding the target. Infer the occluded regions from the geometric cues visible on the target and seamlessly reconstruct its complete 3D structure.
    \item \textbf{Strict preservation of intrinsic attributes:} Faithfully preserve the target's intrinsic material, texture, color distribution, and overall shape characteristics from the original image.
    \item \textbf{Isolation on a pure black background:} Center the complete, unobstructed target object on a pure black (\texttt{\#000000}) background.
    \item \textbf{Neutral-lighting reconstruction:} Remove complex environmental highlights, reflections, and strong cast shadows inherited from the original scene.
\end{enumerate}
Strictly avoid generating any foreground occluders, broken edges, truncated structures, environmental light spots, complex backgrounds, extraneous objects outside the masked target, implausible perspective distortion, blurred textures, or overlays.
\end{quote}

\paragraph{Point sampling.}
We sample $n_{\mathrm{anchor}}=8{,}192$ points from the mesh surface with \texttt{anchor\_from\_geometry}, and sample $n_{\mathrm{target}}=8{,}192$ query points from valid pixels inside the pointmap mask with \texttt{sample\_target\_from\_pointmap}. The RPF input therefore contains 16,384 points. The predicted correspondence positions are read from the query-token segment, implemented as \texttt{pred[n\_anchor:n\_anchor+n\_target]}.

\subsection{RPF training-pair construction and inference details}
\label{sec:appendix_rpf}

\paragraph{Training data and rendering.}
The training set contains approximately $21{,}000\times40$ asset instances, with 20 views rendered for each asset. We use a perspective camera with a 50\,mm focal length, placed 1.5 units along the positive $x$-axis and oriented toward the origin. Asset rotations are sampled uniformly. For each rendered observation, random occlusion is generated using 8 crops before constructing the partial query and its pixel-matched ground-truth correspondence positions. Geometric augmentation applies axis-wise scaling in $[0.9,1.1]$, random scale augmentation in $[0.8,1.2]$, and random translation augmentation in $[0,0.3]$. During training, the entire normal channel is set to zero with probability $0.5$.

\paragraph{Optimization and flow sampling.}
We follow the optimizer, learning-rate, and scheduler settings of the original RPF implementation~\citep{rpf}, and use a batch size of 128. Training uses 2 nodes with 8 NVIDIA A100 GPUs per node. The RPF DiT uses 50 steps in both training and inference.

\paragraph{Closed-form similarity-transform recovery.}
We fit a uniform-scale similarity transform from predicted correspondence positions in the anchor coordinate system to their paired query points in the pointmap scene coordinate system. All correspondences contribute equally; pointmap confidence is not used as a fitting weight. We compute the arithmetic source and target centroids, $\bar{\mathbf{x}}=N^{-1}\sum_i\mathbf{x}_i$ and $\bar{\mathbf{y}}=N^{-1}\sum_i\mathbf{y}_i$, and form the unweighted cross-covariance
\begin{equation}
\mathbf{C}=\mathbf{X}_{c}^{\mathsf T}\mathbf{Y}_{c},
\end{equation}
where $\mathbf{X}_{c}$ and $\mathbf{Y}_{c}$ are the centered source and target points. After singular value decomposition, we enforce a proper rotation by flipping the last row of $\mathbf{V}^{\mathsf T}$ and recomputing $\mathbf{R}$ whenever $\det(\mathbf{R})<0$. The uniform scale and translation are
\begin{equation}
 s=\frac{\sum_i\sigma_i}{\sum_i\lVert\mathbf{x}_{i,c}\rVert_2^2},
 \qquad
 \mathbf{t}=\bar{\mathbf{y}}-s\mathbf{R}\bar{\mathbf{x}}.
\end{equation}
If the scale denominator is at most $10^{-12}$, we set $s=1$. The reported RMSE, mean error, and maximum error are computed from the Euclidean residuals after fitting.

\subsection{Evaluation-time alignment and metric protocol}
\label{sec:appendix_evaluation}
Before computing metrics, we align each predicted scene and its ground-truth counterpart using \texttt{scene\_bbox\_chamfer\_rt}. The coarse bounding-box alignment first fixes the uniform scale as the ratio between the ground-truth and predicted aggregate AABB diagonals, $s=d_{\mathrm{gt}}/d_{\mathrm{pred}}$, and translates the center of the scaled predicted AABB to the ground-truth AABB center. This scale remains fixed thereafter. Starting from the scaled and translated prediction, we use Adam for 60 iterations to optimize only a rigid rotation and translation $(\mathbf{R},\mathbf{t})$ under the Chamfer objective. The resulting single global transform is applied to every predicted object in that scene. Scene-CD, Scene-FScore, and scene-aligned per-object bounding-box IoU are then computed in this common frame. Object-CD and Object-FScore instead use independent object-local normalization followed by FilterReg/ICP rigid registration; they are therefore asset-geometry diagnostics rather than evidence about layout. For the small number of failed scenes or objects, CD is not included in the corresponding CD aggregation, whereas F-score and bounding-box IoU are assigned zero and retained in the average.

\subsection{Complete experiments results}
\label{sec:appendix_additional_experiments}
Tables~\ref{tab:alignment_design_full} and~\ref{tab:provider_replacement_full} provide the complete two-dataset versions of the alignment-design and scene-geometry/object-completion ablations summarized in the main text.

\begin{table*}[t]
\centering
\small
\renewcommand{\arraystretch}{0.8}
\setlength{\tabcolsep}{5pt}
\caption{Complete alignment-design ablation. ``Uncleaned Point'' omits point-cloud cleaning, while ``XYZ only'' removes surface-normal conditioning.}
\label{tab:alignment_design_full}
\begin{tabular}{llccc}
\toprule
Dataset & Variant & Scene CD$\downarrow$ & Scene F$\uparrow$ & BBox IoU$\uparrow$ \\
\midrule
\multirow{3}{*}{Scene-Maker}
& Standard & \textbf{0.242} & \textbf{0.649} & \textbf{0.399} \\
& Uncleaned Point & \textbf{0.242} & 0.646 & \textbf{0.399} \\
& XYZ only & 0.245 & 0.641 & 0.394 \\
\midrule
\multirow{3}{*}{3D-FUTURE200}
& Standard & \textbf{0.140} & \textbf{0.836} & \textbf{0.554} \\
& Uncleaned Point & 0.148 & 0.824 & 0.544 \\
& XYZ only & 0.150 & 0.812 & 0.535 \\
\bottomrule
\end{tabular}
\end{table*}

\begin{table*}[t]
\centering
\small
\renewcommand{\arraystretch}{0.8}
\setlength{\tabcolsep}{4pt}
\caption{Complete scene-geometry and object-completion ablation with Meshy-T2 fixed as the mesh provider and six-channel XYZ--normal RPF input.}
\label{tab:provider_replacement_full}
\begin{tabular}{llccccc}
\toprule
Dataset & Variant & Scene CD$\downarrow$ & Scene F$\uparrow$ & Obj. CD$\downarrow$ & Obj. F$\uparrow$ & BBox IoU$\uparrow$ \\
\midrule
\multirow{3}{*}{Scene-Maker}
& MogeV2 + FLUX-2 & 0.253 & 0.632 & 0.201 & 0.667 & 0.381 \\
& MogeV2 + GPT-Image-2 & 0.254 & 0.630 & 0.208 & 0.656 & 0.381 \\
& DAv3 + FLUX-2 & 0.267 & 0.609 & 0.215 & 0.640 & 0.356 \\
\midrule
\multirow{3}{*}{3D-FUTURE200}
& MogeV2 + FLUX-2 & 0.146 & 0.814 & 0.151 & 0.782 & 0.543 \\
& MogeV2 + GPT-Image-2 & 0.142 & 0.822 & 0.156 & 0.775 & 0.543 \\
& DAv3 + FLUX-2 & 0.169 & 0.772 & 0.173 & 0.743 & 0.476 \\
\bottomrule
\end{tabular}
\end{table*}

\begin{table*}[t]
\centering
\caption{Complete Mesh-provider comparison with all other Easy-3Dfy components fixed. GT provides an upper-bound reference under the current layout workflow.}
\label{tab:mesh_source_ablation_full}

% --- 修改字体命令 ---
\footnotesize % 可选: \footnotesize (适中缩小) 或 \scriptsize (进一步缩小)

\renewcommand{\arraystretch}{0.8}
\setlength{\tabcolsep}{2.0pt}
\begin{tabular}{llccccc}
\toprule
Dataset & Mesh source & Sce. CD$\downarrow$ & Sce. F$\uparrow$ & Obj. CD$\downarrow$ & Obj. F$\uparrow$ & BBox IoU$\uparrow$ \\
\midrule
\multirow{7}{*}{\shortstack{Scene-\\Maker}}
& GT & \textbf{0.242} & \textbf{0.660} & \textbf{0.144} & \textbf{0.789} & \textbf{0.415} \\
& SAM-3Dfy & \textbf{0.242} & 0.649 & 0.188 & 0.695 & 0.399 \\
& TRELLIS-2.0 & 0.284 & 0.557 & 0.273 & 0.548 & 0.317 \\
& HY3D-2.0 & 0.295 & 0.523 & 0.278 & 0.514 & 0.291 \\
& HY3D-3.1 & 0.256 & 0.629 & 0.196 & 0.678 & 0.385 \\
& Meshy-T2 & 0.253 & 0.632 & 0.201 & 0.667 & 0.381 \\
& Tripo-V3 & 0.255 & 0.632 & 0.194 & 0.678 & 0.391 \\
\midrule
\multirow{7}{*}{\shortstack{3D-FUTURE\\200}}
& GT & \textbf{0.111} & \textbf{0.888} & \textbf{0.098} & \textbf{0.878} & \textbf{0.584} \\
& SAM-3Dfy & 0.140 & 0.836 & 0.139 & 0.807 & 0.554 \\
& TRELLIS-2.0 & 0.163 & 0.772 & 0.214 & 0.665 & 0.472 \\
& HY3D-2.0 & 0.245 & 0.626 & 0.261 & 0.553 & 0.388 \\
& HY3D-3.1 & 0.154 & 0.798 & 0.152 & 0.778 & 0.537 \\
& Meshy-T2 & 0.146 & 0.814 & 0.151 & 0.782 & 0.543 \\
& Tripo-V3 & 0.160 & 0.787 & 0.149 & 0.781 & 0.539 \\
\bottomrule
\end{tabular}
\end{table*}

\begin{table*}[t]
\centering
\small
\renewcommand{\arraystretch}{0.8}
\setlength{\tabcolsep}{6pt}
\caption{Effect of RGB conditioning under fixed FLUX-2-pro completion and Meshy-T2 assets. The 9-channel checkpoint is retrained on training pairs that include RGB features. }
\label{tab:rgb_conditioning}
\begin{tabular}{llccc}
\toprule
Dataset & Metric & 9ch: XYZ + normal + RGB & 6ch: XYZ + normal & $\Delta$(6ch$-$9ch) \\
\midrule
\multirow{3}{*}{3D-FUTURE200}
& Scene CD$\downarrow$ & 0.1534 & \textbf{0.1455} & $-0.0079$ \\
& Scene F$\uparrow$ & 0.7973 & \textbf{0.8136} & $+0.0163$ \\
& BBox IoU$\uparrow$ & 0.5231 & \textbf{0.5428} & $+0.0197$ \\
\midrule
\multirow{3}{*}{Scene-Maker}
& Scene CD$\downarrow$ & 0.2617 & \textbf{0.2527} & $-0.0090$ \\
& Scene F$\uparrow$ & 0.6204 & \textbf{0.6320} & $+0.0116$ \\
& BBox IoU$\uparrow$ & 0.3709 & \textbf{0.3812} & $+0.0103$ \\
\bottomrule
\end{tabular}
\end{table*}

Removing RGB improves all reported layout metrics on both datasets under the evaluated setting. This supports the six-channel geometry-only design used in the main experiments: RGB appearance does not provide an additional benefit for the layout-alignment objective here.

\subsection{Additional visualization results}
\label{sup:additional_visualization}
\begin{figure}[H]
\centering
\includegraphics[width=1\textwidth]{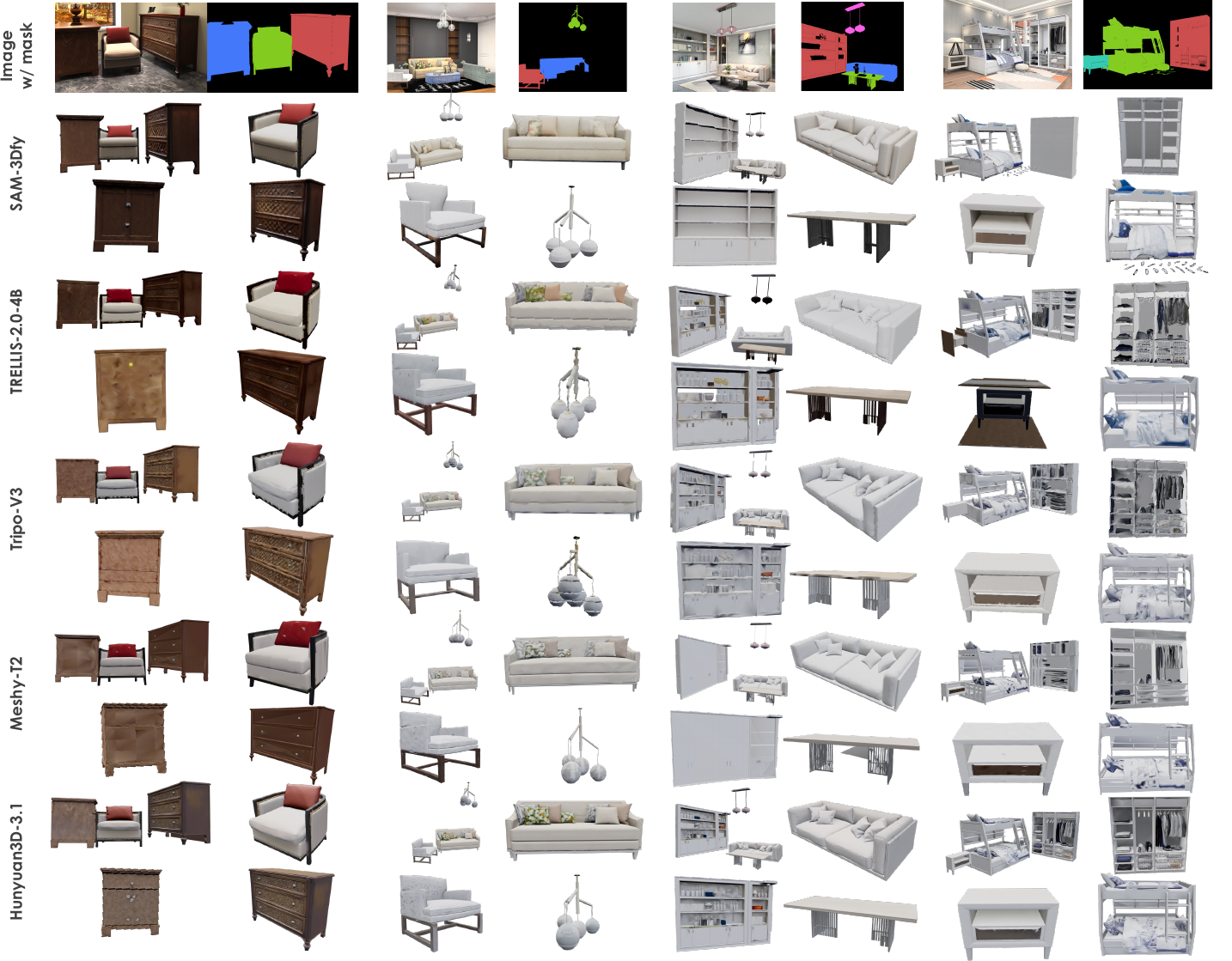}
\caption{\textbf{Additional visualization results on 3D-FUTURE.} 
}
\label{fig:more_vis}
\end{figure}

We provide additional visualization results in the Figure \ref{fig:more_vis}.

\subsection{Inference Efficiency}
The overall inference time of Easy-3Dfy is primarily determined by the runtime of its individual modules and the degree of parallelism. For example, completing a single object using the official FLUX-2-pro API takes on the order of tens of seconds. Low-poly mesh generators such as Meshy-T2 can produce geometries suitable for registration within 10 seconds, whereas more sophisticated models such as Hunyuan3D-V3.1 require approximately 1 minute to generate a complete textured asset. Our RPF-based aligner introduces only a small additional runtime overhead: it supports parallel inference and registration of multiple meshes, with each batch processed within a few seconds.

\subsection{Representative failure cases}
\label{sec:failure_cases}

\begin{figure}[t]
\centering
\includegraphics[width=0.94\textwidth]{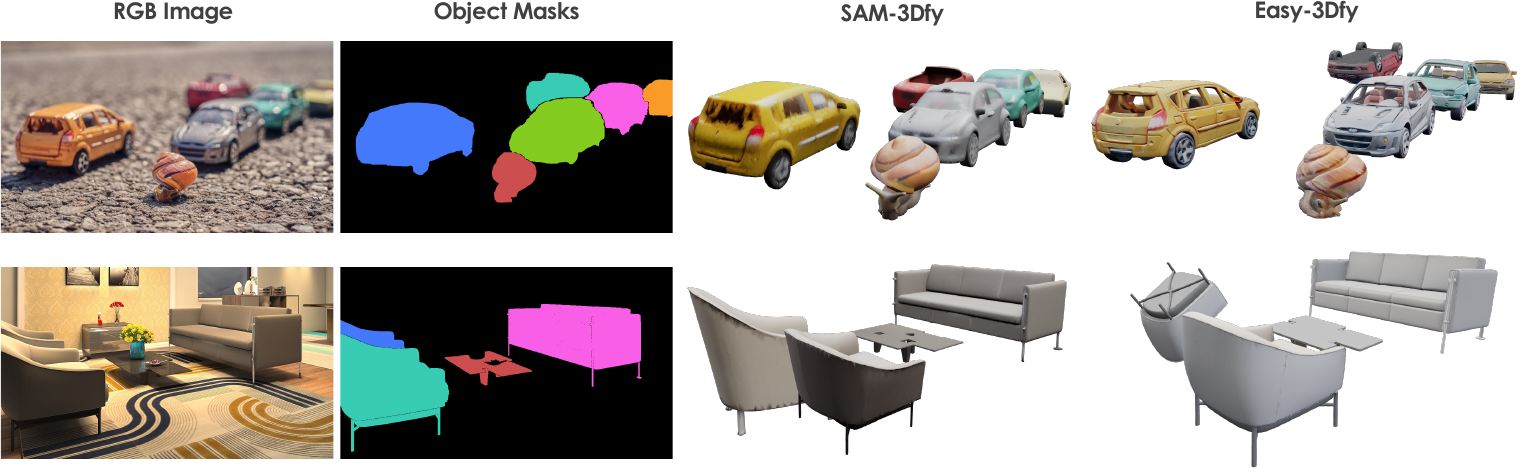}
\caption{\textbf{Failure mode comparison under geometrically challenged conditions.} 
}
\label{fig:failure_cases}
\end{figure}

Layout estimation can easily become underconstrained or fail when object-specific partial geometry is severely degraded or insufficient. Such geometric information loss typically arises from visual challenges, including tiny object scales, severe occlusions, motion or defocus blur, and unreliable depth maps. As illustrated in Figure~\ref{fig:failure_cases}, different paradigms exhibit distinct failure patterns under these challenging scenarios: baseline methods (such as SAM-3D-like frameworks) often suffer from severe drift in scale and translation due to brittle pointmap regression, whereas Easy-3Dfy maintains much better spatial bounded bounds but occasionally experiences rotation instability when geometric constraints become severely underdetermined.

